\documentclass[11pt]{article}
\usepackage{naaclhlt2016}
\usepackage{times}
\usepackage{url}
\usepackage{latexsym}
\usepackage{epsfig}
\usepackage{array}
\usepackage{color}
\usepackage{multirow}
\usepackage{subfig}
\usepackage{algorithmic}
\usepackage{algorithm}
\usepackage{rotating}
\usepackage{booktabs}
\usepackage{makecell}
\usepackage{comment}
\usepackage{amsmath,amsfonts,amssymb}
\usepackage{verbatim}
\usepackage{caption}
\usepackage{booktabs}
\usepackage{ctable}
\usepackage{makecell}
\usepackage{array} 
\usepackage{varwidth} 
\usepackage{tabularx}
\usepackage{enumerate}

\newcommand*{\Scale}[2][4]{\scalebox{#1}{$#2$}}%

\newcolumntype{M}{>{\begin{varwidth}{6cm}}l<{\end{varwidth}}} 
\newcolumntype{C}[1]{>{\centering\let\newline\\\arraybackslash\hspace{0pt}}m{#1}}
\newcolumntype{L}[1]{>{\raggedright\let\newline\\\arraybackslash\hspace{0pt}}m{#1}}
\newcolumntype{Y}{>{\centering\arraybackslash}X}

\naaclfinalcopy 
\def\naaclpaperid{***} 

\title{Multi-domain Neural Network Language Generation for \\ Spoken Dialogue Systems}

\author{Tsung-Hsien Wen, Milica Ga{\v{s}}i\'c, Nikola Mrk{\v{s}}i\'c, Lina M. Rojas-Barahona, \\ {\bf Pei-Hao Su, David Vandyke, Steve Young} \\
  Cambridge University Engineering Department,  \\
  Trumpington Street, 
  Cambridge, CB2 1PZ, UK\\
  {\tt \{thw28,mg436,nm480,lmr46,phs26,djv27,sjy\}@cam.ac.uk}\
}

\date{}

\begin{document}

\maketitle

\begin{abstract}

Moving from limited-domain natural language generation (NLG) to open domain is difficult because the number of semantic input combinations grows exponentially with the number of domains.
Therefore, it is important to leverage existing resources and exploit similarities between domains to facilitate domain adaptation.
In this paper, we propose a procedure to train multi-domain, Recurrent Neural Network-based (RNN) language generators via multiple adaptation steps.  In this procedure,
a model is first trained on counterfeited data synthesised from an out-of-domain dataset, and then fine tuned on a small set of in-domain utterances with a discriminative objective function.
Corpus-based evaluation results show that the proposed procedure can achieve competitive performance in terms of  BLEU score and slot error rate while significantly reducing the data needed to train generators in new, unseen domains.  In subjective testing, human judges confirm that the procedure greatly improves generator performance when only a small amount of data is available in the domain.

\end{abstract}

\section{Introduction}\label{sec:intro}

Modern Spoken Dialogue Systems (SDS) are typically developed according to a well-defined {\it ontology}, which provides a structured representation of the {\it domain} data that the dialogue system can talk about, such as searching for a restaurant or shopping for a laptop.
Unlike conventional approaches employing a substantial amount of handcrafting for each individual processing component~\cite{Ward1994,Bohus2009}, statistical approaches to SDS promise a domain-scalable framework which requires a minimal amount of human intervention~\cite{6407655}. 
\newcite{MrksicSTGSVWY15} showed improved performance in belief tracking by training a general model and adapting it to specific domains.
Similar benefit can be observed in~\newcite{gasiccommittee}, in which a Bayesian committee machine~\cite{Tresp2000} was used to model policy learning in a multi-domain SDS regime.

In past decades, adaptive NLG has been studied from linguistic perspectives, such as systems that learn to tailor user preferences~\cite{Walker07individualand}, convey a specific personality trait~\cite{Mairesse08trainablegeneration,Mairesse2011CUP}, or align with their conversational partner~\cite{Isard06individualityand}.
Domain adaptation was first addressed by~\newcite{Hogan2008} using a generator based on the Lexical Functional Grammar (LFG) f-structures~\cite{KaplanBresnan1082lfg}.
Although these approaches can model rich linguistic phenomenon, they are not readily adaptable to data since they still require many handcrafted rules to define the search space.
Recently, RNN-based language generation has been introduced~\cite{wenrgm15,wensclstm15}.
This class of statistical generators can learn generation decisions directly from dialogue act (DA)-utterance pairs without any semantic annotations~\cite{Mairesse2014} or hand-coded grammars~\cite{Langkilde1998,walker2002training}.
Many existing adaptation approaches~\cite{WenHLTL13,Shi2015,chen2015recurrent} can be directly applied due to the flexibility of the underlying RNN language model (RNNLM) architecture~\cite{39298195}.

Discriminative training (DT) has been successfully used to train RNNs for various tasks.
By optimising directly against the desired objective function such as BLEU score~\cite{export217163} or Word Error Rate ~\cite{Kuo02discriminativetraining}, the model can explore its output space and learn to discriminate between good and bad hypotheses. 
In this paper we show that DT can enable a generator to learn more efficiently when in-domain data is scarce.

The paper presents an incremental recipe for training multi-domain language generators based on a purely data-driven, RNN-based generation model. 
Following a review of related work in section~\ref{sec:related}, 
section~\ref{sec:nlg} describes the detailed RNN generator architecture.
The data counterfeiting approach for synthesising an in-domain dataset is introduced in section~\ref{sec:multi}, where it is compared to the simple model fine-tuning approach.
In section~\ref{sec:xobj}, we describe our proposed DT procedure for training natural language generators.
Following a brief review of the data sets used in section~\ref{sec:dataset},
corpus-based evaluation results are presented in section~\ref{sec:exp}.
In order to assess the subjective performance of our system, a quality test and a pairwise preference test are presented in section~\ref{sec:human}.
The results show that the proposed adaptation recipe improves not only the objective scores but also the user's perceived quality of the system. 
We conclude with a brief summary in section \ref{sec:conclusion}.

\section{Related Work}\label{sec:related}

Domain adaptation problems arise when we have a sufficient amount of labeled data in one domain (the {\it source} domain), but have little or no labeled data in another related domain (the {\it target} domain).
Domain adaptability for real world speech and language applications is especially important because both  language usage and the topics of interest 
are constantly evolving.
Historically, domain adaptation has been less well studied in the NLG community. 
The most relevant work was done by~\newcite{Hogan2008}. They showed that an LFG f-structure based generator could yield better performance when trained on in-domain sentences paired with pseudo parse tree inputs generated from a state-of-the-art, but out-of-domain parser.
The SPoT-based generator proposed by~\newcite{walker2002training} has the potential to address domain adaptation problems. 
However, their published work has focused on tailoring user preferences~\cite{Walker07individualand} and mimicking personality traits~\cite{Mairesse2011CUP}.
~\newcite{adaptivelemon08} proposed a Reinforcement Learning (RL) framework in which policy and NLG components can be jointly optimised and adapted based on online user feedback.
In contrast, ~\newcite{Mairesse2010} has proposed using active learning to mitigate the data sparsity problem when training data-driven NLG systems.
Furthermore, ~\newcite{7078559} trained statistical surface realisers from unlabelled data by an automatic slot labelling technique.

In general, feature-based adaptation is perhaps the most widely used technique~\cite{blitzer2007,Pan2010,DBLP12064660}.
By exploiting correlations and similarities between data points, it has been successfully applied to problems like speaker adaptation~\cite{279278,leggetter95mllr} and various tasks in natural language processing~\cite{DBLPabs09071815}.
In contrast, model-based adaptation is particularly useful for language modeling (LM)~\cite{AMadp_review}. Mixture-based topic LMs~\cite{Gildea99topic} are widely used in N-gram LMs for domain adaptation.
Similar ideas have been applied to  applications that require adapting LMs, such as machine translation (MT)~\cite{Koehn2007EDA} and personalised speech recognition~\cite{wen2012personalized}.

Domain adaptation for Neural Network (NN)-based LMs has also been studied in the past.
A feature augmented RNNLM was first proposed by~\newcite{6424228}, but later applied to multi-genre broadcast speech recognition~\cite{chen2015recurrent} and personalised language modeling~\cite{WenHLTL13}.
These methods are based on fine-tuning existing network parameters on adaptation data. However, careful regularisation is often necessary~\cite{export194346}.
In a slightly different area, ~\newcite{Shi2015} applied curriculum learning to RNNLM adaptation.

Discriminative training (DT)~\cite{Collins2002} is an alternative to the maximum likelihood (ML) criterion. 
For classification, DT can be split into two phases: (1) decoding training examples using the current model and scoring them, and (2) adjusting the model parameters to maximise the separation between the correct target annotation and the competing incorrect annotations.
It has been successfully applied to many research problems, such as speech recognition~\cite{Kuo02discriminativetraining,7178341} and MT~\cite{He2012,export226918}.
Recently,~\newcite{export217163} trained an RNNLM with a DT objective and showed improved performance on an MT task.
However, their RNN probabilities only served as input features to a phrase-based MT system.

\section{The Neural Language Generator}\label{sec:nlg}
The neural language generation model~\cite{wenrgm15,wensclstm15} is a RNNLM~\cite{39298195} augmented 
with semantic input features such as a dialogue act\footnote{A combination of an action type and a set of slot-value pairs. e.g. {\it inform(name="Seven days",food="chinese")}} (DA) denoting the required semantics of the generated output.
At every time step $t$, the model consumes the 1-hot representation of both the DA $\mathrm{\mathbf{d}}_t$ and a token $\mathrm{\mathbf{w}}_{t}$\footnote{
We use {\it token} instead of {\it word} because our model operates on text for which slot values are replaced by their corresponding slot tokens. We call this procedure delexicalisation.} to update its internal state $\mathrm{\mathbf{h}}_t$. Based on this new state, the output distribution over the next output token is calculated.  The model can thus generate a sequence of tokens by repeatedly sampling the current output distribution to obtain the next input token until an end-of-sentence sign is generated.
Finally, the generated sequence is lexicalised\footnote{The process of replacing slot token by its value.} to form the target utterance.

The Semantically Conditioned Long Short-term Memory Network (SC-LSTM)~\cite{wensclstm15} is a specialised extension of the LSTM network~\cite{Hochreiter1997} for language generation which has 
previously been shown capable of learning generation decisions from paired DA-utterances end-to-end without a modular pipeline~\cite{walker2002training,Stent04trainablesentence}.
Like LSTM, SC-LSTM relies on a vector of memory cells $\mathrm{\mathbf{c}}_t \in \mathbb{R}^{n}$ and a set of elementwise multiplication gates to control how information is stored, forgotten, and exploited inside the network.
The SC-LSTM architecture used in this paper is defined by the following equations,
\[
 	\Scale[0.85]{\left(\begin{array}{ll}
		\mathrm{\mathbf{i}}_t\\
		\mathrm{\mathbf{f}}_t\\
		\mathrm{\mathbf{o}}_t\\
		\mathrm{\mathbf{r}}_t\\
		\hat{\mathrm{\mathbf{c}}}_t
	\end{array}\right)
	=
	\left(\begin{array}{ll}
		\text{sigmoid}\\
		\text{sigmoid}\\
		\text{sigmoid}\\
		\text{sigmoid}\\
		\tanh
	\end{array}\right)
	\mathrm{\mathbf{W}}_{5n,2n}
	\left(\begin{array}{ll}
		\mathrm{\mathbf{w}}_t\\
		\mathrm{\mathbf{h}}_{t-1}
	\end{array}\right)}
\]
\[
	\Scale[0.85]{\mathrm{\mathbf{d}}_t = \mathrm{\mathbf{r}}_{t} \odot \mathrm{\mathbf{d}}_{t-1}}
\]
\[
	\Scale[0.85]{\mathrm{\mathbf{c}}_t = \mathrm{\mathbf{f}}_{t} \odot \mathrm{\mathbf{c}}_{t-1} + \mathrm{\mathbf{i}}_{t} \odot \hat{\mathrm{\mathbf{c}}}_t + tanh(\mathrm{\mathbf{W}}_{dc}\mathrm{\mathbf{d}}_t)}
\]
\[
	\Scale[0.85]{\mathrm{\mathbf{h}}_t =  \mathrm{\mathbf{o}}_{t} \odot \tanh(\mathrm{\mathbf{c}}_t)}
\]
where $n$ is the hidden layer size, $\mathrm{\mathbf{i}}_t,\mathrm{\mathbf{f}}_t,\mathrm{\mathbf{o}}_t,\mathrm{\mathbf{r}}_t \in [0,1]^n$ are input, forget, output, and reading gates respectively, $\hat{\mathrm{\mathbf{c}}}_t$ and $\mathrm{\mathbf{c}}_t$ are proposed cell value and true cell value at time $t$,
$\mathrm{\mathbf{W}}_{5n,2n}$ and $\mathrm{\mathbf{W}}_{dc}$ are the model parameters to be learned.
The major difference of the SC-LSTM compared to the vanilla LSTM is the introduction of the reading gates for controlling the semantic input features presented to the network.
It was shown in~\newcite{wensclstm15} that these reading gates act like keyword and key phrase detectors that  learn the alignments between individual semantic input features and their corresponding realisations without additional supervision.

After the hidden layer state is obtained, the computation of the next word distribution and sampling from it is straightforward,
\[
	\Scale[0.85]{p(w_{t+1}|w_t,w_{t-1},...w_0,\mathrm{\mathbf{d}}_t) = softmax( \mathrm{\mathbf{W}}_{ho}\mathrm{\mathbf{h}}_t )}
\]
\[
	\Scale[0.85]{w_{t+1} \sim p(w_{t+1}|w_t,w_{t-1},...w_0,\mathrm{\mathbf{d}}_{t})}.
\]
where $\mathrm{\mathbf{W}}_{ho}$ is another weight matrix to learn.
The entire network is trained end-to-end using a cross entropy cost function, between the predicted word distribution $\mathrm{\mathbf{p}}_t$ and the actual word distribution $\mathrm{\mathbf{y}}_t$, with regularisations on DA transition dynamics,
\begin{equation}
\label{eq:nll}
\Scale[0.85]{F(\theta) = \sum_t\mathrm{\mathbf{p}}_t^{\intercal}log(\mathrm{\mathbf{y}}_t)+ \|\mathrm{\mathbf{d}}_{\text T}\|+\sum_{t=0}^{{\text T}-1}\eta\xi^{\|\mathrm{\mathbf{d}}_{t+1}-\mathrm{\mathbf{d}}_t\|}}
\end{equation}
where $\theta=\{\mathrm{\mathbf{W}}_{5n,2n}, \mathrm{\mathbf{W}}_{dc},\mathrm{\mathbf{W}}_{ho}\}$, $\mathrm{\mathbf{d}}_{\text T}$ is the DA vector at the last index ${\text T}$, and $\eta$ and $\xi$ are constants set to $10^{-4}$ and $100$, respectively. 

\section{Training Multi-domain Models}\label{sec:multi}

Given training instances (represented by DA and sentence tuples $\{d_{i},\Omega_{i}\}$) from the source domain $\mathbb{S}$ (rich) and the target domain $\mathbb{T}$ (limited), the goal is to find a set of SC-LSTM parameters $\theta_{\mathbb{T}}$ that can perform acceptably well in the target domain.

\begin{figure*}[t]
\centerline{\includegraphics[width=150mm]{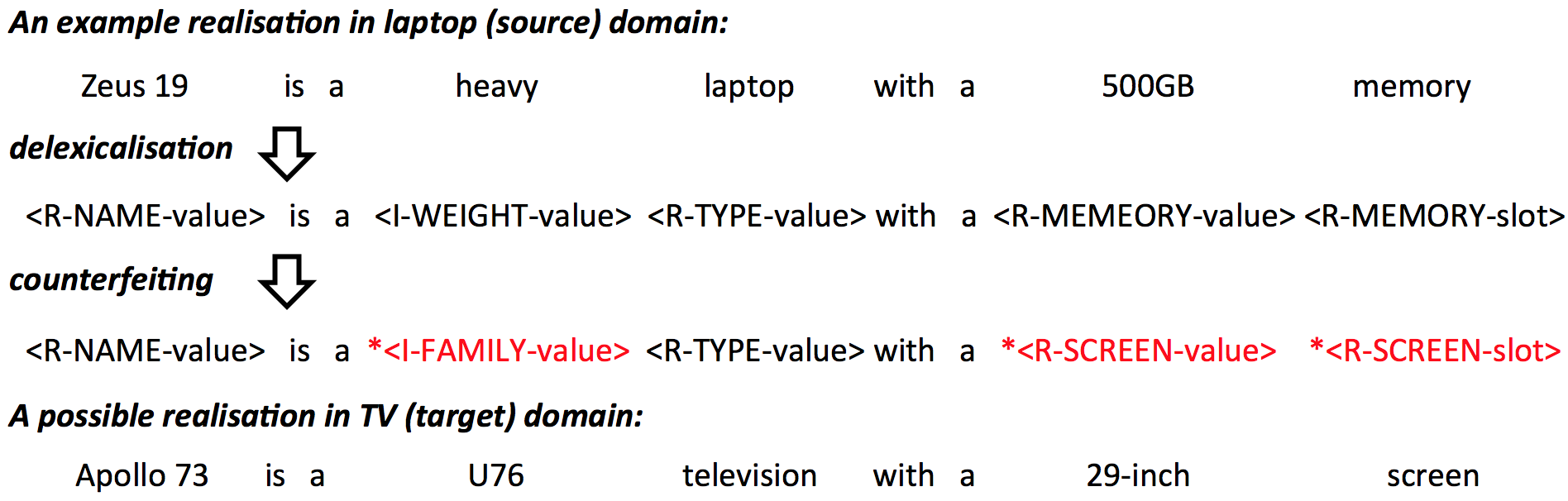}}
\caption{An example of data counterfeiting algorithm. Both slots and values are delexicalised. Slots and values that are not in the target domain are replaced during data counterfeiting (shown in {\it red} with * sign). The prefix inside bracket $<>$ indicates the slot's functional class (I for {\it informable} and R for {\it requestable}). }
\label{fig:counterfeit}
\vspace{-3mm}
\end{figure*}

\subsection{Model Fine-Tuning}\label{ssec:finetune}
A straightforward way to adapt NN-based models to a target domain is to continue training or fine-tuning a well-trained generator on whatever new target domain data is available.
This training procedure is as follows:
\begin{enumerate}
\item Train a source domain generator $\theta_\mathbb{S}$ on source domain data $\{d_{i},\Omega_{i}\}\in \mathbb{S}$ with all values delexicalised\footnote{We have tried training with both slots and values delexicalised and then using the weights to initialise unseen slot-value pairs in the target domain. However, this yielded even worse results since the learned semantic alignment stuck at local minima. Pre-training only the LM parameters did not produce better results either.\label{refnote}}.
\item Divide the adaptation data into training and validation sets. Refine parameters by training on adaptation data $\{d_{i},\Omega_{i}\}\in \mathbb{T}$ with early stopping and a smaller starting learning rate. This yields the target domain generator $\theta_\mathbb{T}$.
\end{enumerate}
Although this method can benefit from parameter sharing of the LM part of the network, the parameters of similar input slot-value pairs are not shared\textsuperscript{\ref{refnote}}.
In other words, realisation of any unseen slot-value pair in the target domain can only be learned from scratch. Adaptation offers no benefit in this case.

\begin{table*}[t]
\vspace{-3mm}
\centering 
\setlength{\intextsep}{3pt plus 2pt minus 2pt} 
\setlength{\abovecaptionskip}{3pt}
\setlength{\belowcaptionskip}{0pt}
\hspace*{0pt}\makebox[\linewidth][c]{%
\scalebox{1.0}{
\begin{tabular}{l|C{6.5cm}|C{6.5cm}}
\hline
		&	Laptop 	&	Television\\
\hline
informable slots		& family, *pricerange, batteryrating, driverange, weightrange, {\bf isforbusinesscomputing}	& family, *pricerange, screensizerange, ecorating, hdmiport, {\bf hasusbport}\\
\hline
requestable slots	& *name, *type, *price, warranty, battery, design, dimension, utility, weight, platform, memory, drive, processor & *name, *type, *price, resolution, powerconsumption, accessories, color, screensize, audio\\
\hline
act type	&\multicolumn{2}{L{13.0cm}}{*inform, *inform\_only\_match, *inform\_on\_match, inform\_all, *inform\_count, inform\_no\_info, *recommend, compare, *select, suggest, *confirm, *request, *request\_more, *goodbye}\\
\hline
\multicolumn{3}{l}{ {\small {\bf bold}=binary slots, *=overlap with SF Restaurant and Hotel domains, all {\it informable slots} can take "dontcare" value}}\\
\end{tabular}}}
\caption{Ontologies for Laptop and TV domains}
\label{tab:otg}
\end{table*}

\subsection{Data Counterfeiting}\label{ssec:forge}

In order to maximise the effect of domain adaptation, the model should be able to (1) generate acceptable realisations for unseen slot-value pairs based on similar slot-value pairs seen in the training data, and (2) continue to distinguish slot-value pairs that are similar but nevertheless distinct.
Instead of exploring weight tying strategies in different training stages (which is complex to implement and typically relies on ad hoc tying rules), we propose instead a data counterfeiting approach to {\it synthesise} target domain data from source domain data.
The procedure is shown in Figure~\ref{fig:counterfeit} and described as following:
\begin{enumerate}
\item Categorise slots in both source and target domain into classes, according to some similarity measure. In our case, we categorise them based on their functional type to yield three classes: {\it informable}, {\it requestable}, and {\it binary}\footnote{{\it Informable} class include all non-binary informable slots while {\it binary} class includes all binary informable slots.}.
\item Delexicalise all slots and values.
\item For each slot $s$ in a source instance $(d_{i},\Omega_{i})\in \mathbb{S}$, randomly select a new slot $s'$ that belongs to both the target ontology and the class of $s$ to replace $s$. Repeat this process for every slot in the instance and yield a new pseudo instance $(\hat{d}_{i},\hat{\Omega}_{i}) \in \mathbb{T}$ in the target domain.
\item Train a generator $\hat{\theta}_\mathbb{T}$ on the counterfeited dataset $\{\hat{d}_{i},\hat{\Omega}_{i}\}\in \mathbb{T}$.
\item Refine parameters on real in-domain data. This yields final model parameters $\theta_\mathbb{T}$.
\end{enumerate}
This approach allows the generator to share realisations among slot-value pairs that have similar functionalities, therefore facilitates the transfer learning of rare slot-value pairs in the target domain.
Furthermore, the approach also preserves the co-occurrence statistics of slot-value pairs and their realisations. 
This allows the model to learn the gating mechanism even before adaptation data is introduced.

\section{Discriminative Training}\label{sec:xobj}

In contrast to the traditional ML criteria (Equation~\ref{eq:nll}) whose goal is to maximise the log-likelihood of correct examples, DT aims at separating correct examples from competing incorrect examples. 
Given a training instance $(d_{i},\Omega_{i})$,
the training process starts by generating a set of candidate sentences $Gen(d_i)$ using the current model parameter $\theta$ and DA $d_i$. 
The discriminative cost function can therefore be written as 
\begin{align}
\label{eq:xobj}
F(\theta) &= -\mathbb{E}[L(\theta)] \nonumber \\
& = -\sum_{\Omega\in Gen(d_i)} p_{\theta}(\Omega|d_{i}) L(\Omega,\Omega_{i})
\end{align}
where $L(\Omega,\Omega_{i})$ is the scoring function evaluating candidate $\Omega$ by taking ground truth $\Omega_{i}$ as reference.
$p_{\theta}(\Omega|d_{i})$ is the normalised probability of the candidate and is calculated by
\begin{equation}
\label{eq:prnn}
\Scale[1.15]{p_{\theta}(\Omega|d_i) = \frac{\textrm{exp}[\gamma \log p(\Omega|d_i,\theta)]}   {\sum_{\Omega'\in Gen(d_i)}\textrm{exp}[\gamma \log p(\Omega'|d_i,\theta)]}}
\end{equation}
$\gamma \in [0,\infty]$ is a tuned scaling factor that flattens the distribution for $\gamma<1$ and sharpens it for $\gamma>1$.
The unnormalised candidate likelihood $\log p(\Omega|d_i,\theta)$ is produced by summing token likelihoods from the RNN generator output,
\begin{equation}
\label{eq:prnn}
\log p(\Omega|d_i,\theta) = \sum_{w_{t}\in \Omega}\log p(w_t|d_i,\theta)
\end{equation}

The scoring function $L(\Omega,\Omega_{i})$ can be further generalised to take several scoring functions into account
\begin{equation}
L(\Omega,\Omega_{i}) = \sum_{j} L_{j}(\Omega,\Omega_{i}) \beta_{j}
\end{equation}
where $\beta_j$ is the weight for $j$-th scoring function.
Since the cost function presented here (Equation~\ref{eq:xobj}) is differentiable everywhere, back propagation can be applied to calculate the gradients and update parameters directly. 

\begin{figure}[t]
    \vspace{-2mm}
    \centering
    \subfloat[BLEU score curve]{{\includegraphics[width=7.7cm]{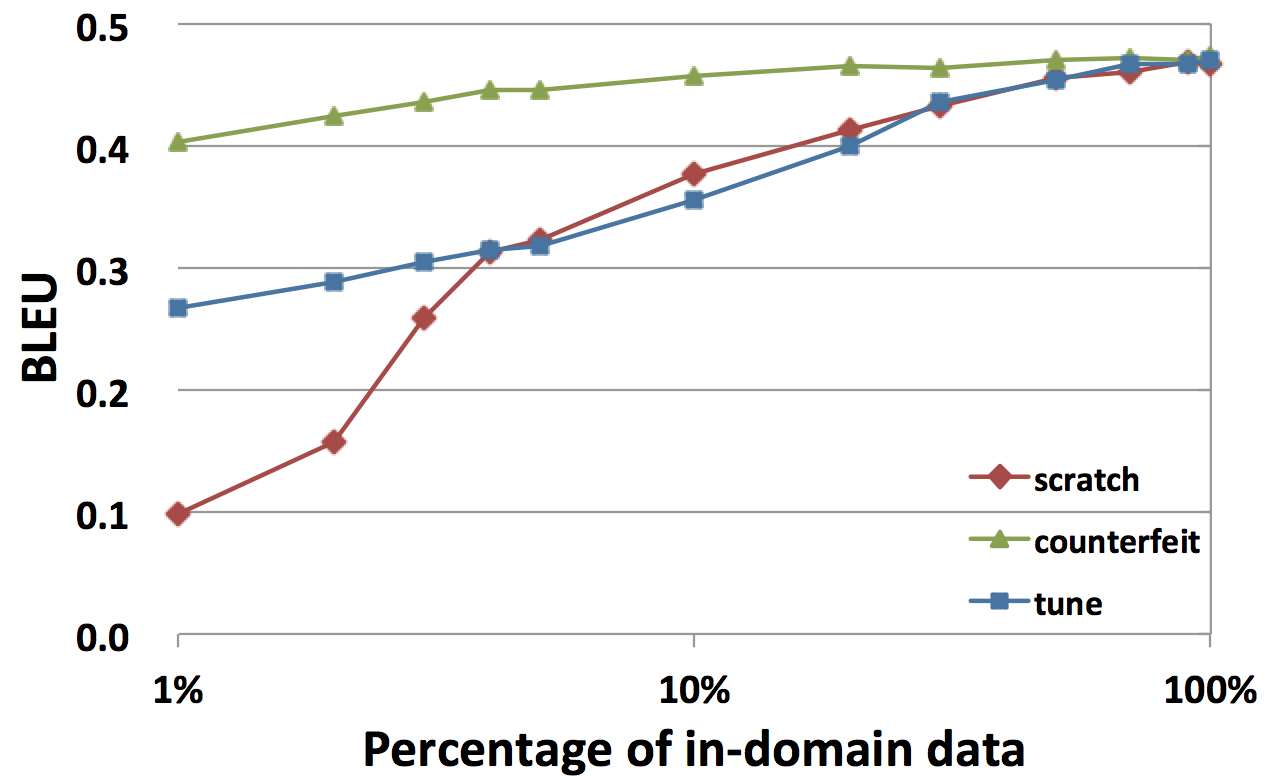} }}
    \qquad
    \subfloat[Slot error rate curve]{{\includegraphics[width=7.7cm]{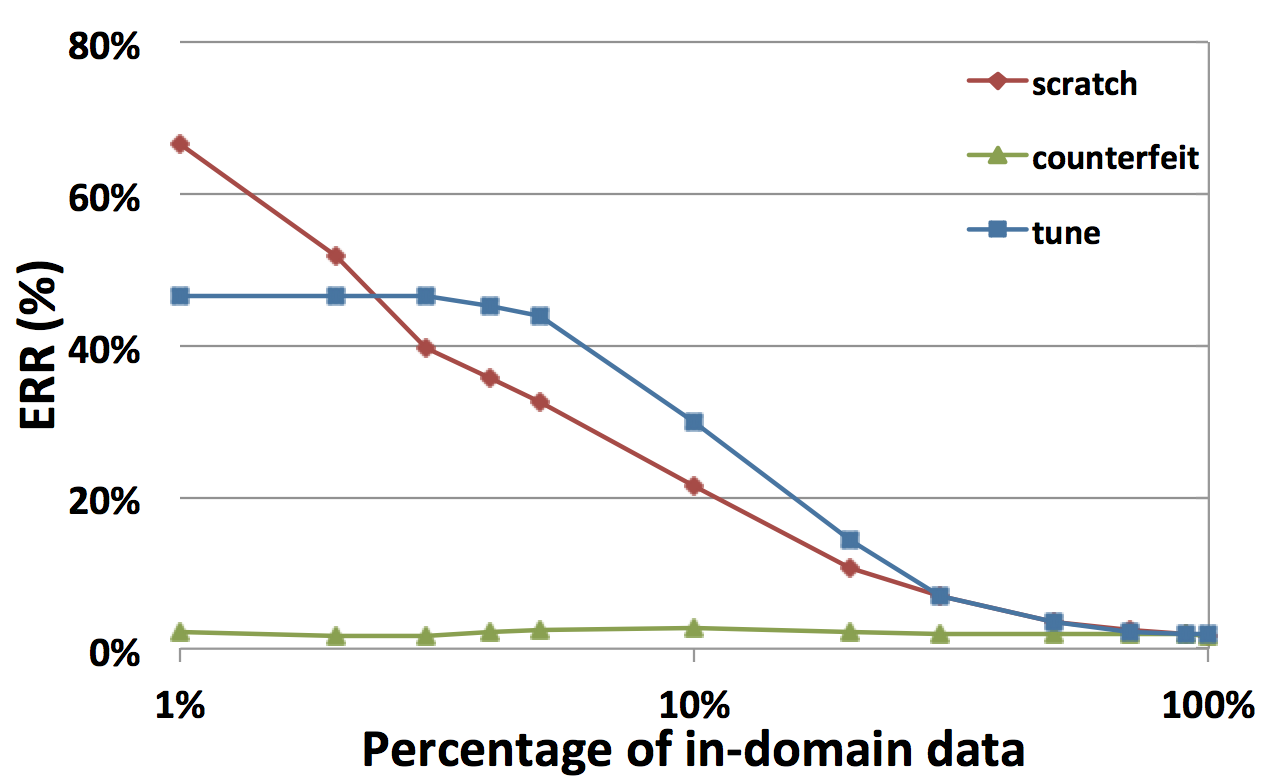} }}
    \caption{Results evaluated on TV domain by adapting models from laptop domain. Comparing train-from-scratch model ({\it scratch}) with model fine-tuning approach ({\it tune}) and data counterfeiting method ({\it counterfeit}). $10\%\approx 700$ examples.}
    \label{fig:l2t}
        \vspace{-3mm}
\end{figure}

\begin{figure}[t]
    \vspace{-2mm}
    \centering
    \subfloat[BLEU score curve]{{\includegraphics[width=7.7cm]{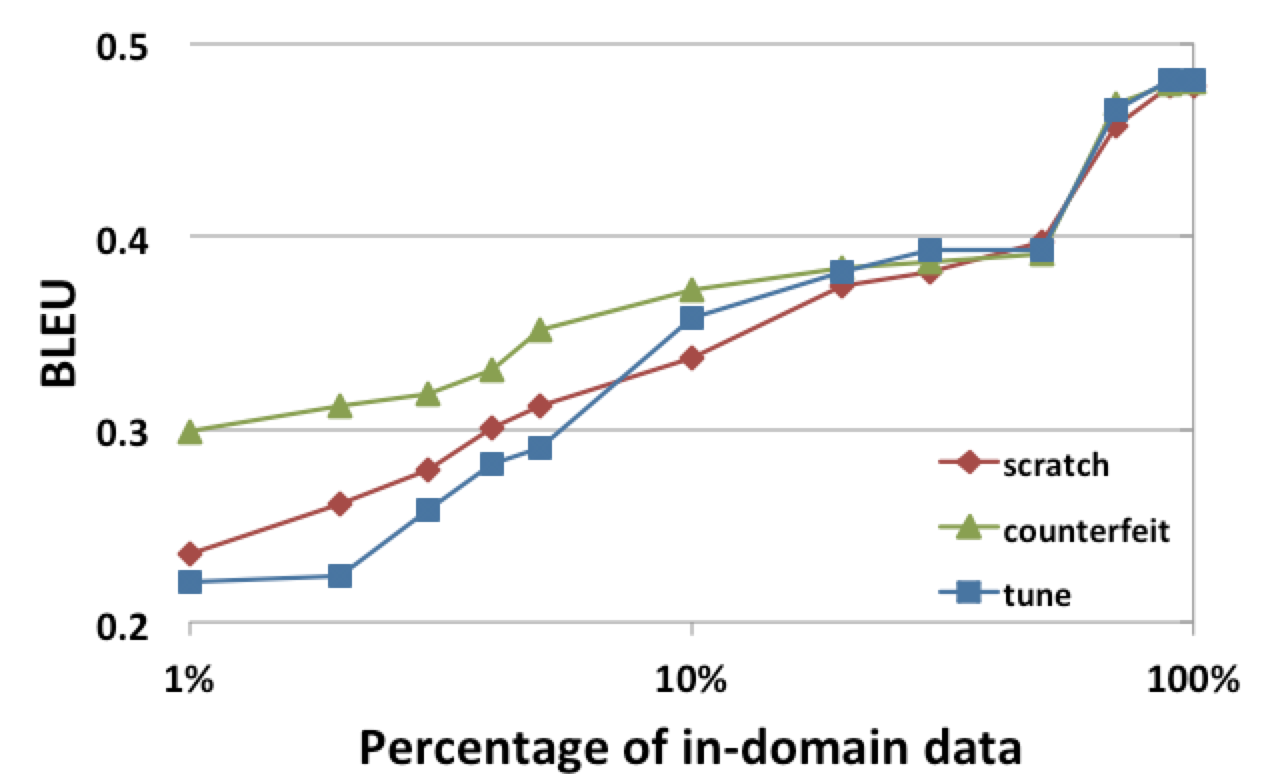} }}
    \qquad
    \subfloat[Slot error rate curve]{{\includegraphics[width=7.7cm]{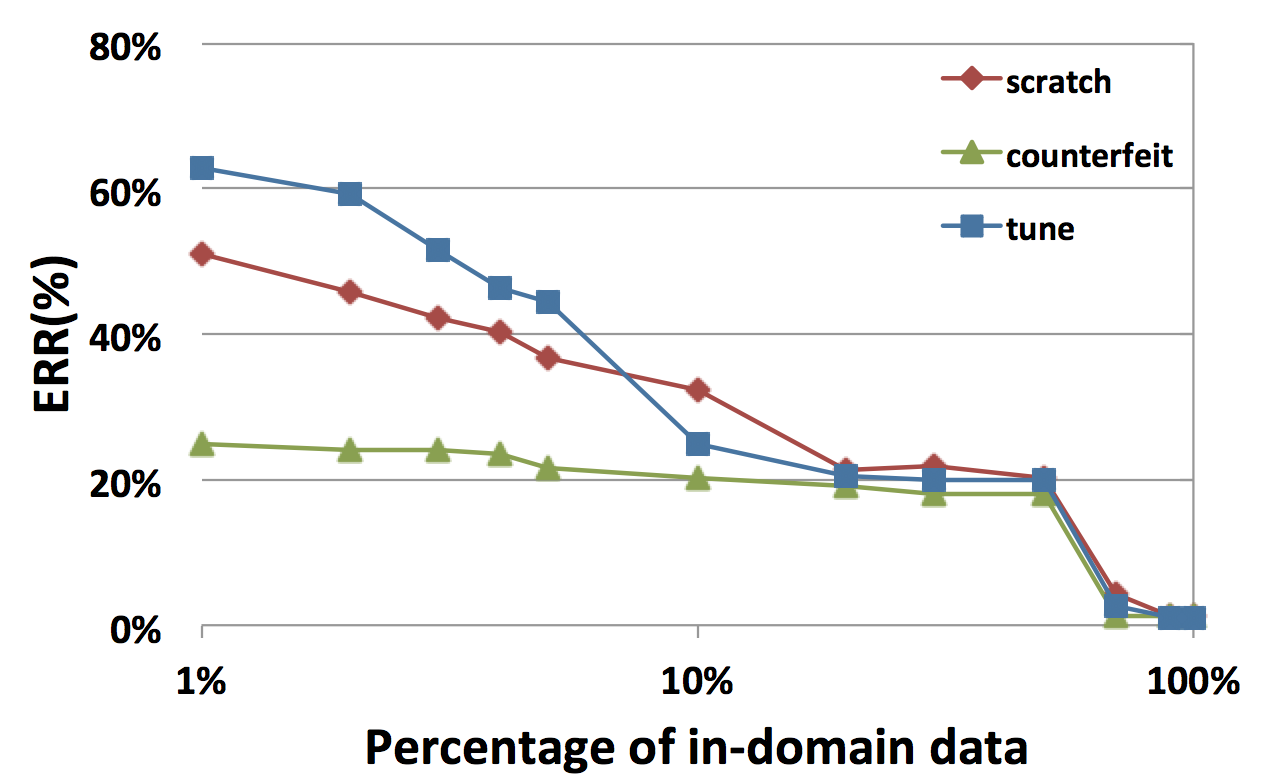} }}
    \caption{The same set of comparison as in Figure~\ref{fig:l2t}, but the results were evaluated by adapting from SF restaurant and hotel joint dataset to laptop and TV joint dataset. $10\%\approx 2K$ examples.}
    \label{fig:rh2lt}
    \vspace{-3mm}
\end{figure}

\section{Datasets}\label{sec:dataset}

In order to test our proposed recipe for training multi-domain language generators, we conducted experiments using four different domains: finding a restaurant, finding a hotel, buying a laptop, and buying a television. 
Datasets for the restaurant and hotel domains have been previously released by~\newcite{wensclstm15}.  These were created by workers recruited by Amazon Mechanical Turk (AMT) by asking them to propose an appropriate natural language realisation corresponding to each system dialogue act actually generated by a dialogue system.  However, the number of actually occurring DA combinations in the restaurant and hotel domains were rather limited ($\sim$200) and since multiple references were collected for each DA, the resulting datasets are not sufficiently diverse to enable the assessment of the generalisation capability of the different training methods over unseen semantic inputs.

In order to create more diverse datasets for the laptop and TV domains, we enumerated all possible combinations of dialogue act types and slots based on the ontology shown in Table~\ref{tab:otg}.
This yielded about 13K distinct DAs in the  laptop domain and 7K distinct DAs in the TV domain.
We then used AMT workers to collect just one realisation for each DA.
Since the resulting datasets have a much larger input space but only one training example for each DA, the system must learn partial realisations of concepts and be able to recombine and apply them to unseen DAs.
Also note that the number of act types and slots of the new ontology is larger, which makes NLG in both laptop and TV domains much harder.

\section{Corpus-based Evaluation}\label{sec:exp}

We first assess generator performance using two objective evaluation metrics, the BLEU-4 score~\cite{papineni2002bleu} and slot error rate $\mathrm{ERR}$~\cite{wensclstm15}.
Slot error rates were calculated by averaging slot errors over each of the top 5 realisations in the entire corpus. 
We used multiple references to compute the BLEU scores when available (i.e. for the restaurant and hotel domains).
In order to better compare results across different methods, we plotted the BLEU and slot error rate curves against different amounts of adaptation data.
Note that in the graphs the {\it x}-axis is presented on a log-scale.

\subsection{Experimental Setup}\label{ssec:setup}

The generators were implemented using the Theano library \cite{bergstra2010,Theano2012}, and trained by partitioning each of the collected corpora into a training, validation, and testing set in the ratio 3:1:1.
All the generators were trained by treating each sentence as a mini-batch.
An $l_2$ regularisation term was added to the objective function for every 10 training examples.
The hidden layer size was set to be 100 for all cases.
Stochastic gradient descent and back propagation through time \cite{werbos1990backpropagation} were used to optimise the parameters.
In order to prevent overfitting, early stopping was implemented using the validation set.

During decoding, we over-generated 20 utterances and selected the top 5 realisations for each DA according to the following reranking criteria,
\begin{equation}
\label{eq:score}
\Scale[0.85]{R = - ( F(\theta) + \lambda \mathrm{ERR} )}
\end{equation}
where $\lambda$ is a tradeoff constant, $F(\theta)$ is the cost generated by network parameters $\theta$, and the slot error rate $\mathrm{ERR}$ is computed by exact matching of the slot tokens in the candidate utterances.
$\lambda$ is set to a large value (10) in order to severely penalise nonsensical outputs.
Since our generator works stochastically and the trained networks can differ depending on the initialisation, all the results shown below were averaged over 5 randomly initialised networks.

\subsection{Data Counterfeiting}\label{ssec:adapt}

We first compared the data counterfeiting ({\it counterfeit}) approach with the model fine-tuning ({\it tune}) method and models trained from scratch ({\it scratch}). 
Figure~\ref{fig:l2t} shows the result of adapting models between similar domains, from laptop to TV.
Because of the parameter sharing in the LM part of the network, model fine-tuning ({\it tune}) achieves a better BLEU score than training from scratch ({\it scratch}) when target domain data is limited.
However, if we apply the data counterfeiting ({\it counterfeit}) method, we obtain an even greater BLEU score gain.
This is mainly due to the better realisation of unseen slot-value pairs.
On the other hand, data counterfeiting ({\it counterfeit}) also brings a substantial reduction in slot error rate.
This is because it preserves the co-occurrence statistics between slot-value pairs and realisations, which allows the model to learn good semantic alignments even before adaptation data is introduced.
Similar results can be seen in Figure~\ref{fig:rh2lt}, in which adaptation was performed on more disjoint domains: restaurant and hotel joint domain to laptop and TV joint domain.
The data counterfeiting ({\it counterfeit}) method is still superior to the other methods. 

\begin{figure}[t]
    \vspace{-2mm}
    \centering
    \subfloat[Effect of DT on BLEU]{{\includegraphics[width=7.7cm]{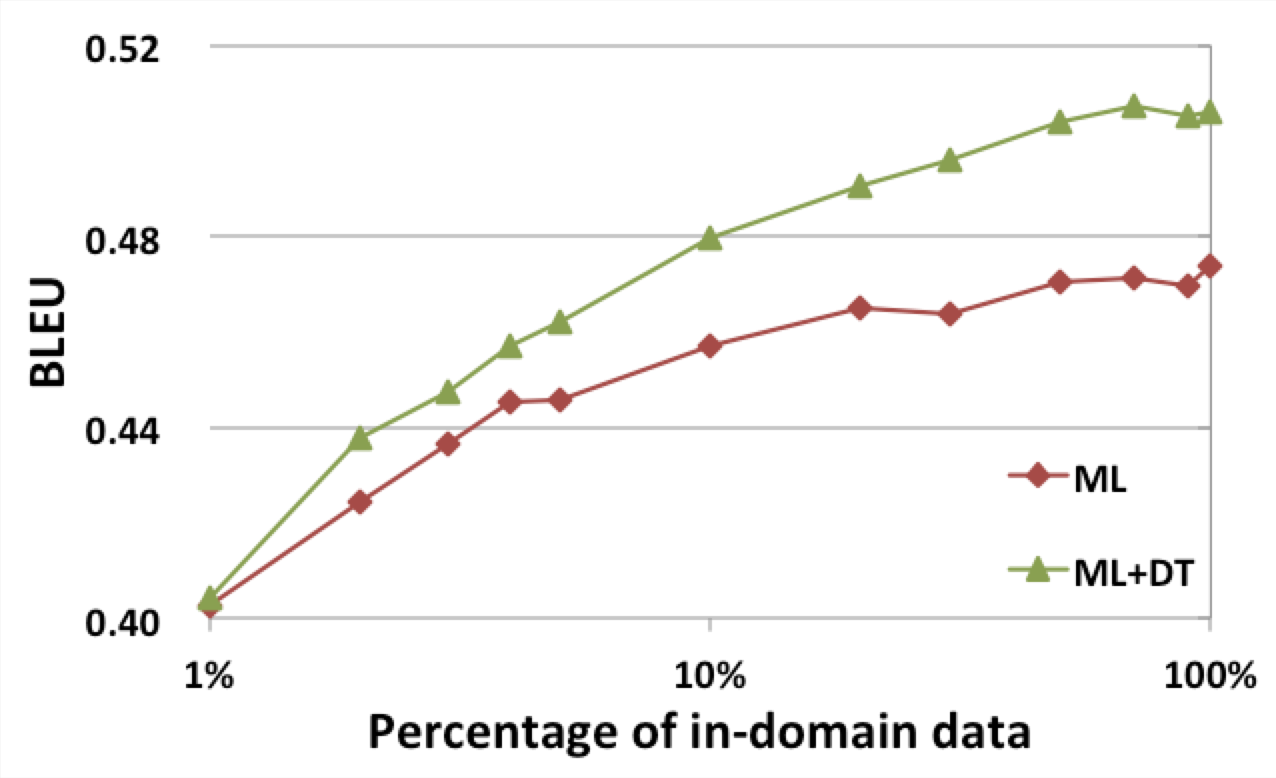} }}
    \qquad
    \subfloat[Effect of DT on slot error rate]{{\includegraphics[width=7.7cm]{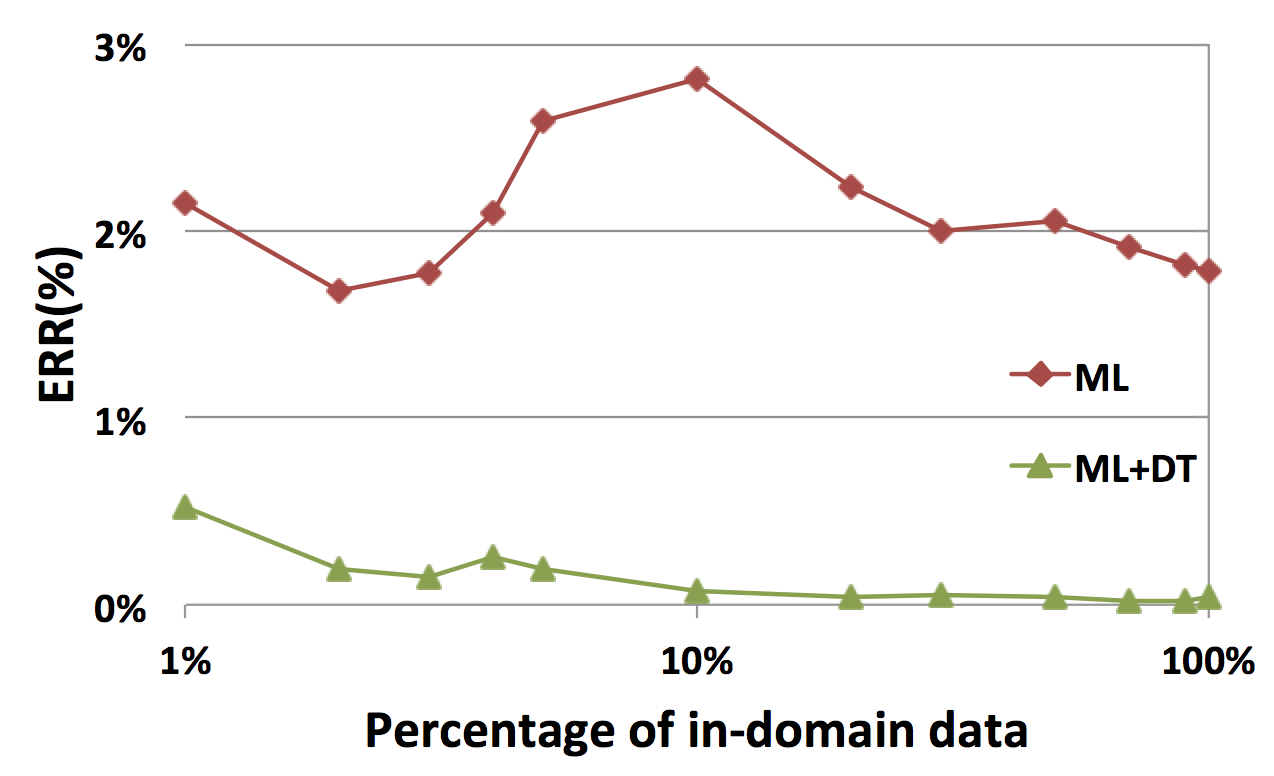} }}
    \caption{Effect of applying DT training after ML adaptation. The results were evaluated on laptop to TV adaptation. $10\%\approx 700$ examples.}
    \label{fig:xobj}
    \vspace{-3mm}
\end{figure}

\subsection{Discriminative Training}\label{ssec:xobj}

The generator parameters obtained from data counterfeiting and ML adaptation were further tuned by applying DT. 
In each case, the models were optimised using two objective  functions: BLEU-4 score and slot error rate.
However, we used a soft version of BLEU called sentence BLEU as described in~\newcite{export217163}, to mitigate the sparse n-gram match problem of BLEU at the sentence level.
In our experiments, we set $\gamma$ to 5.0 and $\beta_{j}$ to 1.0 and -1.0 for BLEU and ERR, respectively.
For each DA, we applied our generator 50 times to generate candidate sentences. Repeated candidates were removed.
We treated the remaining candidates as a single batch and updated the model parameters by the procedure described in section~\ref{sec:xobj}.
We evaluated performance of the algorithm on the laptop to TV adaptation scenario, and compared models with and without discriminative training ({\it ML+DT} \& {\it ML}). 
The results are shown in Figure~\ref{fig:xobj} where it can be seen that  DT consistently improves generator performance on both metrics. 
Another interesting point to note is that slot error rate is easier to optimise compared to BLEU (ERR$\rightarrow0$ after DT).  This is probably because the sentence BLEU optimisation criterion is only an approximation of the corpus BLEU score used for evaluation.

\begin{table}[t]
\centering 
\setlength{\intextsep}{3pt plus 2pt minus 2pt} 
\setlength{\abovecaptionskip}{2pt}
\setlength{\belowcaptionskip}{0pt}
\setlength\extrarowheight{1pt}
\hspace*{0pt}\makebox[\linewidth][c]{%
\scalebox{0.98}{
\begin{tabularx}{\columnwidth}{@{}l|YY||YY@{}}
\Xhline{2\arrayrulewidth}
\multirow{2}{*}{{\bf Method}} & \multicolumn{2}{c||}{{\bf TV to Laptop}} & \multicolumn{2}{c}{{\bf laptop to TV}}\\
						& {\bf Info.}& {\bf Nat.} 	& {\bf Info.}	& {\bf Nat.}\\ \cline{2-5}
\Xhline{2\arrayrulewidth}
scrALL   					&	2.64		   	 	&      2.37  			&	2.54				&	2.36\\
\Xhline{1\arrayrulewidth}
DT-10\%	       	&	{\bf 2.52\tmark[{\makebox[0pt][l]{**}}]}&	{\bf2.25\tmark[{\makebox[0pt][l]{**}}]}	&	{\bf2.51}	&	2.19\tmark[{\makebox[0pt][l]{**}}]\\
ML-10\%    	&	2.51\tmark[{\makebox[0pt][l]{**}}]   	&      2.22\tmark[{\makebox[0pt][l]{**}}]	&	2.45\tmark[{\makebox[0pt][l]{**}}]	&	{\bf2.22}\tmark[{\makebox[0pt][l]{**}}]\\
\Xhline{1\arrayrulewidth}
scr-10\%	       	&	2.24\tmark[{\makebox[0pt][l]{**}}]	&	2.03\tmark[{\makebox[0pt][l]{**}}]	&	2.00\tmark[{\makebox[0pt][l]{**}}]	&	1.92\tmark[{\makebox[0pt][l]{**}}]\\
\Xhline{2\arrayrulewidth}
\multicolumn{5}{l}{* p \textless 0.05, ** p \textless 0.005}
\end{tabularx}
}}
\caption{Human evaluation for utterance quality in two domains. Results are shown in two metrics (rating out of 3). Statistical significance was computed using a two-tailed Student's t-test, between the model trained with full data ({\it scrALL}) and all others. }
\label{tab:compare}
\vspace{-3mm}
\end{table}

\section{Human Evaluation}\label{sec:human}

Since automatic metrics may not consistently agree with human perception \cite{Stent05evaluatingevaluation}, human testing is needed to assess subjective quality.
To do this, a set of judges were recruited using AMT. 
We tested our models on two adaptation scenarios: laptop to TV and TV to laptop.
For each task, two systems among the four were compared: training from scratch using full dataset ({\it scrALL}), adapting with DT training but only 10\% of target domain data ({\it DT-10\%}), adapting with ML training but only 10\% of target domain data ({\it ML-10\%}), and training from scratch using only 10\% of target domain data ({\it scr-10\%}). 
In order to evaluate system performance in the presence of language variation, each system generated 5 different surface realisations for each input DA and the human judges were asked to score each of them in terms of informativeness and naturalness (rating out of 3), and also asked to state a preference between the two.
Here {\it informativeness} is defined as whether the utterance contains all the information specified in the DA, and {\it naturalness} is defined as whether the utterance could plausibly have been produced by a human.
In order to decrease the amount of information presented to the judges, utterances that appeared identically in both systems were filtered out.
We tested about 2000 DAs for each scenario distributed uniformly between contrasts except that allowed 50\% more comparisons between {\it ML-10\%} and {\it DT-10\%} because they were close.

Table~\ref{tab:compare} shows the subjective quality assessments which exhibit the same general trend as the objective results.
If a large amount of target domain data is available, training everything from scratch ({\it scrALL}) achieves a very good performance and adaptation is not necessary. 
However, if only a limited amount of in-domain data is available, efficient adaptation is critical ({\it DT-10\%} \& {\it ML-10\%} $>$ {\it scr-10\%}).
Moreover,  judges also preferred the DT trained generator ({\it DT-10\%}) compared to the ML trained generator ({\it ML-10\%}), especially for {\it informativeness}.
In the laptop to TV scenario, the {\it informativeness} score of DT method ({\it DT-10\%}) was considered indistinguishable when comparing to the method trained with full training set ({\it scrALL}). 
The preference test results are shown in Table \ref{tab:pref}.
Again, adaptation methods ({\it DT-10\%} \& {\it ML-10\%}) are crucial to bridge the gap between domains when the target domain data is scarce ({\it DT-10\%} \& {\it ML-10\%} $>$ {\it scr-10\%}).
The results also suggest that the DT training approach ({\it DT-10\%}) was preferred compared to ML training ({\it ML-10\%}), even though the preference in this case was not statistically significant.

\begin{table}[t]
  \centering
  \setlength\extrarowheight{1pt}
  \subfloat[Preference test on TV to laptop adaptation scenario]{%
  \scalebox{0.88}{
    \begin{tabular}{l|cccc}
        \Xhline{2\arrayrulewidth}
        {\bf Pref.\%}	&	{\bf scr-10\%}	&	{\bf ML-10\%}	&	{\bf DT-10\%}	&	{\bf scrALL}\\
        \Xhline{1\arrayrulewidth}
        {\bf scr-10\%}	&	-		&	34.5\tmark[{\makebox[0pt][l]{**}}]		&	33.9\tmark[{\makebox[0pt][l]{**}}]	& 	22.4\tmark[{\makebox[0pt][l]{**}}]\\
	{\bf ML-10\%}	&	65.5\tmark[{\makebox[0pt][l]{**}}]		&	-		&	44.9		&	36.8\tmark[{\makebox[0pt][l]{**}}]\\
	{\bf DT-10\%}	&	66.1\tmark[{\makebox[0pt][l]{**}}] 	&	55.1		&	-		&	35.9\tmark[{\makebox[0pt][l]{**}}]\\
	{\bf scrALL}	&	77.6\tmark[{\makebox[0pt][l]{**}}]	&	63.2\tmark[{\makebox[0pt][l]{**}}]	&	64.1\tmark[{\makebox[0pt][l]{**}}]		&	-\\
        \Xhline{2\arrayrulewidth}
        \multicolumn{5}{l}{* p \textless 0.05, ** p \textless 0.005}
    \end{tabular}}
    \hspace{.5cm}%
  }\hspace{1cm}\vspace{.2mm}
  \subfloat[Preference test on laptop to TV adaptation scenario]{%
    \scalebox{0.88}{
    \begin{tabular}{l|cccc}
        \Xhline{2\arrayrulewidth}
        {\bf Pref.\%}	&	{\bf scr-10\%}	&	{\bf ML-10\%}	&	{\bf DT-10\%}	&	{\bf scrALL}\\
        \Xhline{1\arrayrulewidth}
        {\bf scr-10\%}	&	-		&	17.4\tmark[{\makebox[0pt][l]{**}}]		&	14.2\tmark[{\makebox[0pt][l]{**}}]	& 	14.8\tmark[{\makebox[0pt][l]{**}}]\\
	{\bf ML-10\%}	&	82.6\tmark[{\makebox[0pt][l]{**}}]		&	-		&	48.1		&	37.1\tmark[{\makebox[0pt][l]{**}}]\\
	{\bf DT-10\%}	&	85.8\tmark[{\makebox[0pt][l]{**}}] 	&	51.9		&	-		&	41.6\tmark[{\makebox[0pt][l]{*}}]\\
	{\bf scrALL}	&	85.2\tmark[{\makebox[0pt][l]{**}}]	&	62.9\tmark[{\makebox[0pt][l]{**}}]	&	58.4\tmark[{\makebox[0pt][l]{*}}]		&	-\\
        \Xhline{2\arrayrulewidth}
        \multicolumn{5}{l}{* p \textless 0.05, ** p \textless 0.005}
    \end{tabular}}
    \hspace{.5cm}%
  }
   \caption{Pairwise preference test among four approaches in two domains. Statistical significance was computed using two-tailed binomial test.}
  \label{tab:pref}
  \vspace{-3mm}
\end{table}

\section{Conclusion and Future Work}\label{sec:conclusion}

In this paper we have proposed a procedure for training multi-domain, RNN-based language generators, by data counterfeiting and discriminative training.
The procedure is general and applicable to any data-driven language generator. Both corpus-based evaluation and human assessment were performed.
Objective measures on corpus data have demonstrated that by applying this procedure to adapt models between four different dialogue domains,  good performance can be achieved with much less training data.
Subjective assessment by human judges confirm the effectiveness of the approach.

The proposed domain adaptation method requires a small amount of annotated data to be collected offline. In our future work, we intend to focus on training the generator on the fly with real user feedback during conversation.

\section*{Acknowledgments}
Tsung-Hsien Wen and David Vandyke are supported by Toshiba Research Europe Ltd, Cambridge Research Laboratory.

\bibliographystyle{naaclhlt2016}
\bibliography{naaclhlt2016}

\begin{thebibliography}{}

\bibitem[\protect\citename{Auli and Gao}2014]{export217163}
Michael Auli and Jianfeng Gao.
\newblock 2014.
\newblock Decoder integration and expected bleu training for recurrent neural
  network language models.
\newblock In {\em Proceedings of ACL}. Association for Computational
  Linguistics.

\bibitem[\protect\citename{Auli \bgroup et al.\egroup }2014]{export226918}
Michael Auli, Michel Galley, and Jianfeng Gao.
\newblock 2014.
\newblock Large-scale expected bleu training of phrase-based reordering models.
\newblock In {\em Proceedings of EMNLP}. Association for Computational
  Linguistics.

\bibitem[\protect\citename{Bastien \bgroup et al.\egroup }2012]{Theano2012}
Fr{\'{e}}d{\'{e}}ric Bastien, Pascal Lamblin, Razvan Pascanu, James Bergstra,
  Ian~J. Goodfellow, Arnaud Bergeron, Nicolas Bouchard, and Yoshua Bengio.
\newblock 2012.
\newblock Theano: new features and speed improvements.
\newblock Deep Learning and Unsupervised Feature Learning NIPS 2012 Workshop.

\bibitem[\protect\citename{Bellegarda}2004]{AMadp_review}
Jerome~R. Bellegarda.
\newblock 2004.
\newblock Statistical language model adaptation: review and perspectives.
\newblock {\em Speech Communication}.

\bibitem[\protect\citename{Bergstra \bgroup et al.\egroup }2010]{bergstra2010}
James Bergstra, Olivier Breuleux, Fr{\'{e}}d{\'{e}}ric Bastien, Pascal Lamblin,
  Razvan Pascanu, Guillaume Desjardins, Joseph Turian, David Warde-Farley, and
  Yoshua Bengio.
\newblock 2010.
\newblock Theano: a {CPU} and {GPU} math expression compiler.
\newblock In {\em Proceedings of the Python for Scientific Computing
  Conference}.

\bibitem[\protect\citename{Blitzer \bgroup et al.\egroup }2007]{blitzer2007}
John Blitzer, Mark Dredze, and Fernando Pereira.
\newblock 2007.
\newblock Biographies, bollywood, boom-boxes and blenders: Domain adaptation
  for sentiment classification.
\newblock In {\em Proceedings of ACL}. Association for Computational
  Linguistics.

\bibitem[\protect\citename{Bohus and Rudnicky}2009]{Bohus2009}
Dan Bohus and Alexander~I. Rudnicky.
\newblock 2009.
\newblock The ravenclaw dialog management framework: Architecture and systems.
\newblock {\em Computer Speech and Language}.

\bibitem[\protect\citename{Chen \bgroup et al.\egroup }2015]{chen2015recurrent}
Xie Chen, Tan Tian, Liu Xunying, Lanchantin Pierre, Wan Moquan, Mark Gales, and
  Woodland Phil.
\newblock 2015.
\newblock Recurrent neural network language model adaptation for multi-genre
  broadcast speech recognition.
\newblock In {\em Proceedings of InterSpeech}.

\bibitem[\protect\citename{Collins}2002]{Collins2002}
Michael Collins.
\newblock 2002.
\newblock Discriminative training methods for hidden markov models: Theory and
  experiments with perceptron algorithms.
\newblock In {\em Proceedings of EMNLP}. Association for Computational
  Linguistics.

\bibitem[\protect\citename{Cuayáhuitl \bgroup et al.\egroup }2014]{7078559}
H.~Cuayáhuitl, N.~Dethlefs, H.~Hastie, and X.~Liu.
\newblock 2014.
\newblock Training a statistical surface realiser from automatic slot
  labelling.
\newblock In {\em Spoken Language Technology Workshop (SLT), 2014 IEEE}, pages
  112--117, Dec.

\bibitem[\protect\citename{Daum{\'{e}}~III}2009]{DBLPabs09071815}
Hal Daum{\'{e}}~III.
\newblock 2009.
\newblock Frustratingly easy domain adaptation.
\newblock {\em CoRR}, abs/0907.1815.

\bibitem[\protect\citename{Duan \bgroup et al.\egroup }2012]{DBLP12064660}
Lixin Duan, Dong Xu, and Ivor~W. Tsang.
\newblock 2012.
\newblock Learning with augmented features for heterogeneous domain adaptation.
\newblock {\em CoRR}, abs/1206.4660.

\bibitem[\protect\citename{Ga{\v{s}}i\'c \bgroup et al.\egroup
  }2015]{gasiccommittee}
Milica Ga{\v{s}}i\'c, Nikola Mrk{\v{s}}i\'c, Pei{-}hao Su, David Vandyke,
  Tsung{-}Hsien Wen, and Steve~J. Young.
\newblock 2015.
\newblock Policy committee for adaptation in multi-domain spoken dialogue
  systems.
\newblock In {\em Proceedings of ASRU}.

\bibitem[\protect\citename{Gauvain and Lee}1994]{279278}
Jean-Luc Gauvain and Chin-Hui Lee.
\newblock 1994.
\newblock Maximum a posteriori estimation for multivariate gaussian mixture
  observations of markov chains.
\newblock {\em Speech and Audio Processing, IEEE Transactions on}.

\bibitem[\protect\citename{Gildea and Hofmann}1999]{Gildea99topic}
Daniel Gildea and Thomas Hofmann.
\newblock 1999.
\newblock Topic-based language models using em.
\newblock In {\em Proceedings of EuroSpeech}.

\bibitem[\protect\citename{He and Deng}2012]{He2012}
Xiaodong He and Li~Deng.
\newblock 2012.
\newblock Maximum expected bleu training of phrase and lexicon translation
  models.
\newblock In {\em Proceedings of ACL}. Association for Computational
  Linguistics.

\bibitem[\protect\citename{Hochreiter and Schmidhuber}1997]{Hochreiter1997}
Sepp Hochreiter and J\"{u}rgen Schmidhuber.
\newblock 1997.
\newblock Long short-term memory.
\newblock {\em Neural Computation}.

\bibitem[\protect\citename{Hogan \bgroup et al.\egroup }2008]{Hogan2008}
Deirdre Hogan, Jennifer Foster, Joachim Wagner, and Josef van Genabith.
\newblock 2008.
\newblock Parser-based retraining for domain adaptation of probabilistic
  generators.
\newblock In {\em Proceedings of INLG}. Association for Computational
  Linguistics.

\bibitem[\protect\citename{Isard \bgroup et al.\egroup
  }2006]{Isard06individualityand}
Amy Isard, Carsten Brockmann, and Jon Oberlander.
\newblock 2006.
\newblock Individuality and alignment in generated dialogues.
\newblock In {\em Proceedings of INLG}. Association for Computational
  Linguistics.

\bibitem[\protect\citename{Kaplan and Bresnan}1982]{KaplanBresnan1082lfg}
Ronald~M. Kaplan and Joan Bresnan.
\newblock 1982.
\newblock Lexical-{F}unctional {G}rammar: a formal system for grammatical
  representation.
\newblock In Joan Bresnan, editor, {\em The mental representation of
  grammatical relations}. MIT Press.

\bibitem[\protect\citename{Koehn and Schroeder}2007]{Koehn2007EDA}
Philipp Koehn and Josh Schroeder.
\newblock 2007.
\newblock Experiments in domain adaptation for statistical machine translation.
\newblock In {\em Proceedings of StatMT}. Association for Computational
  Linguistics.

\bibitem[\protect\citename{Kuo \bgroup et al.\egroup
  }2002]{Kuo02discriminativetraining}
Hong-kwang Kuo, Eric Fosler-lussier, Hui Jiang, and Chin-hui Lee.
\newblock 2002.
\newblock Discriminative training of language models for speech recognition.
\newblock In {\em Proceedings of ICASSP}.

\bibitem[\protect\citename{Langkilde and Knight}1998]{Langkilde1998}
Irene Langkilde and Kevin Knight.
\newblock 1998.
\newblock Generation that exploits corpus-based statistical knowledge.
\newblock In {\em Proceedings of ACL}. Association for Computational
  Linguistics.

\bibitem[\protect\citename{Leggetter and Woodland}1995]{leggetter95mllr}
Chris Leggetter and Philip Woodland.
\newblock 1995.
\newblock Maximum likelihood linear regression for speaker adaptation of
  continuous density hidden {Markov} models.
\newblock {\em Computer Speech and Language}.

\bibitem[\protect\citename{Lemon}2008]{adaptivelemon08}
Oliver Lemon.
\newblock 2008.
\newblock Adaptive natural language generation in dialogue using reinforcement
  learning.
\newblock In {\em Proceedings of SemDial}.

\bibitem[\protect\citename{Mairesse and
  Walker}2008]{Mairesse08trainablegeneration}
François Mairesse and Marilyn Walker.
\newblock 2008.
\newblock Trainable generation of big-five personality styles through
  data-driven parameter estimation.
\newblock In {\em Proceedings of ACL}. Association for Computational
  Linguistics.

\bibitem[\protect\citename{Mairesse and Walker}2011]{Mairesse2011CUP}
Fran\c{c}ois Mairesse and Marilyn~A. Walker.
\newblock 2011.
\newblock Controlling user perceptions of linguistic style: Trainable
  generation of personality traits.
\newblock {\em Computer Linguistics}.

\bibitem[\protect\citename{Mairesse and Young}2014]{Mairesse2014}
Fran\c{c}ois Mairesse and Steve Young.
\newblock 2014.
\newblock Stochastic language generation in dialogue using factored language
  models.
\newblock {\em Computer Linguistics}.

\bibitem[\protect\citename{Mairesse \bgroup et al.\egroup }2010]{Mairesse2010}
Fran\c{c}ois Mairesse, Milica Ga\v{s}i\'{c}, Filip Jur\v{c}\'{\i}\v{c}ek, Simon
  Keizer, Blaise Thomson, Kai Yu, and Steve Young.
\newblock 2010.
\newblock Phrase-based statistical language generation using graphical models
  and active learning.
\newblock In {\em Proceedings of ACL}. Association for Computational
  Linguistics.

\bibitem[\protect\citename{Mikolov and Zweig}2012]{6424228}
Tom\'a{\v{s}} Mikolov and Geoffrey Zweig.
\newblock 2012.
\newblock Context dependent recurrent neural network language model.
\newblock In {\em Proceedings of SLT}.

\bibitem[\protect\citename{Mikolov \bgroup et al.\egroup }2010]{39298195}
Tom\'a{\v{s}} Mikolov, Martin Karafi√°t, Luk\'a{\v{s}} Burget, Jan
  {\v{C}}ernock\'y, and Sanjeev Khudanpur.
\newblock 2010.
\newblock {Recurrent neural network based language model}.
\newblock In {\em Proceedings of InterSpeech}.

\bibitem[\protect\citename{Mrk{\v{s}}i\'c \bgroup et al.\egroup
  }2015]{MrksicSTGSVWY15}
Nikola Mrk{\v{s}}i\'c, Diarmuid~{\'{O}} S{\'{e}}aghdha, Blaise Thomson, Milica
  Ga{\v{s}}i\'c, Pei{-}hao Su, David Vandyke, Tsung{-}Hsien Wen, and Steve~J.
  Young.
\newblock 2015.
\newblock Multi-domain dialog state tracking using recurrent neural networks.
\newblock {\em CoRR}, abs/1506.07190.

\bibitem[\protect\citename{Pan and Yang}2010]{Pan2010}
Sinno~Jialin Pan and Qiang Yang.
\newblock 2010.
\newblock A survey on transfer learning.
\newblock {\em IEEE Trans. on Knowledge and Data Engineering}.

\bibitem[\protect\citename{Papineni \bgroup et al.\egroup
  }2002]{papineni2002bleu}
Kishore Papineni, Salim Roukos, Todd Ward, and Wei-Jing Zhu.
\newblock 2002.
\newblock Bleu: a method for automatic evaluation of machine translation.
\newblock In {\em Proceedings of ACL}. Association for Computational
  Linguistics.

\bibitem[\protect\citename{Shi \bgroup et al.\egroup }2015]{Shi2015}
Yangyang Shi, Martha Larson, and Catholijn~M. Jonker.
\newblock 2015.
\newblock Recurrent neural network language model adaptation with curriculum
  learning.
\newblock {\em Computer, Speech and Language}.

\bibitem[\protect\citename{Stent \bgroup et al.\egroup
  }2004]{Stent04trainablesentence}
Amanda Stent, Rashmi Prasad, and Marilyn Walker.
\newblock 2004.
\newblock Trainable sentence planning for complex information presentation in
  spoken dialog systems.
\newblock In {\em Proceedings of ACL}. Association for Computational
  Linguistics.

\bibitem[\protect\citename{Stent \bgroup et al.\egroup
  }2005]{Stent05evaluatingevaluation}
Amanda Stent, Matthew Marge, and Mohit Singhai.
\newblock 2005.
\newblock Evaluating evaluation methods for generation in the presence of
  variation.
\newblock In {\em Proceedings of CICLing 2005}.

\bibitem[\protect\citename{Tresp}2000]{Tresp2000}
Volker Tresp.
\newblock 2000.
\newblock A bayesian committee machine.
\newblock {\em Neural Computation}.

\bibitem[\protect\citename{Voigtlaender \bgroup et al.\egroup }2015]{7178341}
P.~Voigtlaender, P.~Doetsch, S.~Wiesler, R.~Schluter, and H.~Ney.
\newblock 2015.
\newblock Sequence-discriminative training of recurrent neural networks.
\newblock In {\em Proceedings of ICASSP}.

\bibitem[\protect\citename{Walker \bgroup et al.\egroup
  }2002]{walker2002training}
Marilyn~A Walker, Owen~C Rambow, and Monica Rogati.
\newblock 2002.
\newblock Training a sentence planner for spoken dialogue using boosting.
\newblock {\em Computer Speech and Language}.

\bibitem[\protect\citename{Walker \bgroup et al.\egroup
  }2007]{Walker07individualand}
Marilyn Walker, Amanda Stent, FranÁois Mairesse, and Rashmi Prasad.
\newblock 2007.
\newblock Individual and domain adaptation in sentence planning for dialogue.
\newblock {\em Journal of Artificial Intelligence Research (JAIR)}.

\bibitem[\protect\citename{Ward and Issar}1994]{Ward1994}
Wayne Ward and Sunil Issar.
\newblock 1994.
\newblock Recent improvements in the cmu spoken language understanding system.
\newblock In {\em Proceedings of Workshop on HLT}. Association for
  Computational Linguistics.

\bibitem[\protect\citename{Wen \bgroup et al.\egroup
  }2012]{wen2012personalized}
Tsung-Hsien Wen, Hung-Yi Lee, Tai-Yuan Chen, and Lin-Shan Lee.
\newblock 2012.
\newblock Personalized language modeling by crowd sourcing with social network
  data for voice access of cloud applications.
\newblock In {\em Proceedings of SLT}.

\bibitem[\protect\citename{Wen \bgroup et al.\egroup }2013]{WenHLTL13}
Tsung-Hsien Wen, Aaron Heidel, Hung yi~Lee, Yu~Tsao, and Lin-Shan Lee.
\newblock 2013.
\newblock Recurrent neural network based language model personalization by
  social network crowdsourcing.
\newblock In {\em Proceedings of InterSpeech}.

\bibitem[\protect\citename{Wen \bgroup et al.\egroup }2015a]{wenrgm15}
Tsung-Hsien Wen, Milica Ga{\v{s}}i\'c, Dongho Kim, Nikola Mrk{\v{s}}i\'c,
  Pei-Hao Su, David Vandyke, and Steve Young.
\newblock 2015a.
\newblock Stochastic language generation in dialogue using recurrent neural
  networks with convolutional sentence reranking.
\newblock In {\em Proceedings of SIGdial}. Association for Computational
  Linguistics.

\bibitem[\protect\citename{Wen \bgroup et al.\egroup }2015b]{wensclstm15}
Tsung-Hsien Wen, Milica Ga{\v{s}}i\'c, Nikola Mrk{\v{s}}i\'c, Pei-Hao Su, David
  Vandyke, and Steve Young.
\newblock 2015b.
\newblock Semantically conditioned lstm-based natural language generation for
  spoken dialogue systems.
\newblock In {\em Proceedings of EMNLP}. Association for Computational
  Linguistics.

\bibitem[\protect\citename{Werbos}1990]{werbos1990backpropagation}
Paul~J Werbos.
\newblock 1990.
\newblock Backpropagation through time: what it does and how to do it.
\newblock {\em Proceedings of the IEEE}.

\bibitem[\protect\citename{Young \bgroup et al.\egroup }2013]{6407655}
Steve Young, Milica Ga{\v{s}}i\'c, Blaise Thomson, and Jason~D. Williams.
\newblock 2013.
\newblock Pomdp-based statistical spoken dialog systems: A review.
\newblock {\em Proceedings of the IEEE}.

\bibitem[\protect\citename{Yu \bgroup et al.\egroup }2013]{export194346}
Dong Yu, Kaisheng Yao, Hang Su, Gang Li, and Frank Seide.
\newblock 2013.
\newblock Kl-divergence regularized deep neural network adaptation for improved
  large vocabulary speech recognition.
\newblock In {\em Proceedings of ICASSP}.

\end{thebibliography}

\end{document}